\documentclass[conference]{IEEEtran}
\IEEEoverridecommandlockouts
\usepackage{cite}
\usepackage{float}
\usepackage{amsmath,amssymb,amsfonts}
\usepackage{algorithmic}
\usepackage{subcaption}
\usepackage[ruled,lined,linesnumbered,commentsnumbered,boxed]{algorithm2e}
\usepackage{graphicx}
\usepackage{textcomp}
\usepackage{xcolor}
\usepackage{mathtools}
\usepackage{multirow}
\usepackage{hyperref}
\usepackage{array}
\usepackage{comment}
\usepackage{physics}
\def\BibTeX{{\rm B\kern-.05em{\sc i\kern-.025em b}\kern-.08em
    T\kern-.1667em\lower.7ex\hbox{E}\kern-.125emX}}
    
\makeatletter
\def\footnoterule{\kern-3\p@
  \hrule \@width 2in \kern 2.6\p@} 
\makeatother    

\makeatletter
\newcommand{\linebreakand}{%
  \end{@IEEEauthorhalign}
  \hfill\mbox{}\par
  \mbox{}\hfill\begin{@IEEEauthorhalign}
}
\makeatother

\newcommand{\bigoh}[1]{\mathcal{O}(#1)}

\hypersetup{
    colorlinks=true,
    citecolor=black,
    linkcolor=black,
    filecolor=magenta,      
    urlcolor=blue,
}

\begin{document}

\title{A Swarm Variant for the Schrödinger Solver
}

\author{
\IEEEauthorblockN{1\textsuperscript{st} Urvil Nileshbhai Jivani}
\IEEEauthorblockA{\textit{CSIS} \\
\textit{BITS Pilani K K Birla Goa Campus}\\
Goa, India \\
f20170943@goa.bits-pilani.ac.in}
\and
\IEEEauthorblockN{2\textsuperscript{nd} Omatharv Bharat Vaidya}
\IEEEauthorblockA{\textit{CSIS \& Mathematics} \\
\textit{BITS Pilani K K Birla Goa Campus}\\
Goa, India \\
f20180354@goa.bits-pilani.ac.in}
\and
\IEEEauthorblockN{3\textsuperscript{rd} Anwesh Bhattacharya}
\IEEEauthorblockA{\textit{CSIS \& Physics} \\
\textit{BITS Pilani}\\
Pilani, India \\
f2016590@pilani.bits-pilani.ac.in}
\linebreakand
\IEEEauthorblockN{4\textsuperscript{th} Snehanshu Saha}
\IEEEauthorblockA{\textit{CSIS \& APPCAIR} \\
\textit{BITS Pilani K K Birla Goa Campus}\\
Goa, India \\
snehanshu.saha@ieee.org}
}

\IEEEoverridecommandlockouts
\IEEEpubid{\makebox[\columnwidth]{978-1-5386-5541-2/18/\$31.00~\copyright2018 IEEE \hfill} \hspace{\columnsep}\makebox[\columnwidth]{ }}
\maketitle
\IEEEpubidadjcol

\begin{abstract}
This paper introduces the application of the Exponentially Averaged Momentum Particle Swarm Optimization (EM-PSO) as a derivative-free optimizer for Neural Networks. It adopts PSO's major advantages such as search space exploration and higher robustness to local minima compared to gradient-descent optimizers such as Adam. Neural network based solvers endowed with gradient optimization are now being used to approximate solutions to Differential Equations. Here, we demonstrate the novelty of EM-PSO in approximating gradients and leveraging the property in solving the Schrödinger equation, for the Particle-in-a-Box problem. We also provide the optimal set of hyper-parameters supported by mathematical proofs, suited for our algorithm\footnote{Communicating Author: Snehanshu Saha}.
\end{abstract}
\section{Introduction}
The Schrödinger Equation \cite{griffiths} is the central equation in quantum mechanics describing the evolution of a particle, or in general, an abstract quantum state. We solve the one-dimensional time-independent version of it by modelling it as a neural-network optimization problem to be optimized by a variant of a Particle Swarm Optimizer, EM-PSO \cite{saha2020adaswarn} ---
\begin{gather}
    -\frac{\hbar^2}{2m}\dv[2]{\psi(x)}{x} + V(x)\psi(x) = E\psi(x) \label{eq:uni.schrodinger}
\end{gather}
In the past few decades, Neural Networks (NN) have become quite popular due to their accuracy and efficacy in solving difficult problems. Their contribution to Data mining, Brain-Machine Interface, Pattern Recognition, Cyber-security, Bio-medical engineering, etc has been monumental. We intend to exploit NN to solve the Schrödinger equation. Instead of the frequently used Gradient-descent algorithm for back-propagation, we use a novel meta-heuristic optimization algorithm EM-PSO. Our methods used for the Schrödinger equation could potentially be applied to differential equations in general.

We begin with the motivation for EM-PSO. Section \ref{sec:schrodinger} presents the Schrödinger equation. Then, section \ref{NN} gives an introduction to the theory of NN and their utilization to solve Differential equations. Further, section \ref{sec:contribution} describes the working of EM-PSO in detail. Section \ref{results} presents the experiment we did using the developed theory and section \ref{summary} states the conclusion.

\section{Background \& Motivation} 

The number of potentials in quantum mechanics for which analytically exact solution exist is very few, and has been well noted in \cite{griffiths, sakurai}. Typical potentials are usually solvable in terms of the hypergeometric, and confluent hypergeometric functions, such as the Woods-Saxon potential used in modelling effective inter-nuclear forces \cite{sqrt.pot}. Multi-atomic/multi-electronic systems are beyond the realm of obtaining analytical solutions, even for the simples cases of the Hydrogen molecule or the Helium atom \cite{bransden}. In the context of multi-atomic systems of Quantum Chemistry, numerous techniques exist to approximate the wavefunction  --- \cite{symbolic.quantum} and references therein --- that depend on tailoring Physics concepts to code such as the Variational Principle, Ritz Ansatz, etc with additional constraints to model a particular problem. It would desirable to have a generic wavefunction solver suitable for a variety of potentials. Our work is an effort in this direction, and we introduce novel techniques such as probability regularization (Section \ref{5.5}), that would work in any given potential. The loss function thus created is computationally expensive for Adam to optimize (\emph{gradient computation}), and hence we use Adaswarm\cite{saha2020adaswarn} to tackle it.

Several works have shown the efficiency of gradient-dependent NN in solving differential equations \cite{neurodiffeq}. In 1999, \textit{Lagaris et. al.} \cite{lagaris1998ann} used NN to solve differential equations by considering a trial solution that satisfies initial and boundary conditions. \textit{NeuroDiffEq} \cite{neurodiffeq} is a modern python implementation of such a technique. A recent work \cite{manoj2020nonn} investigated the advantages of employing an extreme machine learning algorithm for obtaining the optimal values of NN weights over the numerical optimization of the loss function. \cite{a.malek} combined the approaches of feed-forward NN and numerical optimization to obtain a Hybrid method, that inherited the benefits of both these techniques. \cite{cedric.Flamant} proposed that a NN be used as a solution bundle, a collection of solutions to an ODE for various initial states and system parameters.  \cite{magill.Qureshi} introduced a technique based on the singular vector canonical correlation analysis (SVCCA) and illustrated this method on NNs trained to solve parametrized BVPs from Poisson PDE. \cite{craig.milos} presented a novel framework for solving irregular PDEs using Deep NN (DNNs). \cite{piscopo.waite} proposed a novel way of
finding a numerical solution to wide classes of differential equations without using trial solutions. \cite{yifan.sun} came up with a NN based approach for extracting models from dynamic data using ODEs and PDEs.

Some papers have worked on the application of such techniques to essential problems. For instance, \cite{tim.dockhorn} extensively explored the applications of solving the Poisson and the steady Navier–Stokes equations using NNs. \cite{guo.cao} introduced a Physics Informed Neural Network (PINN) method to solve PDEs. 

Even though some argue in favor of classical analytic methods to solve differential equations arising in Physics [CITE], there are caveats to sticking to such rigid approaches \cite{duch.geerd}. The approximation by discretization is tedious, computationally expensive, and that there is no guarantee of convergence to the analytical solution. The process for the construction of the trial function could be avoided by training the NN to satisfy initial/boundary conditions along with the differential equation optimization function.  This serves as the motivation to solve a challenging problem using gradient-free optimization.

We realized that in order to expand the problems that could be optimized, we need to use gradient-free optimizers. For instance, problems involving non-differentiable functions, discrete feasible space, large dimensionality, multiple local minima could be difficult to solve using standard gradient-based back-propagation methods like gradient descent. A 2015 study, \cite{alick2015nogradient} proposed a gradient-free numerical optimization-based control scheme to solve the problem of formation control and target tracking in multi-agent systems. This method helped overlook the strong assumption that the gradient or the Hessian of the objective function could be analytically computed from continuously measured system states.

In this paper, we explore the use of EM-PSO, another gradient-free optimizer that has more flexibility than vanilla PSO and overcomes the problem of stagnation in local minima. It was also well established that EM-PSO supplements the exploration part of PSO by giving more weight to the exploration part, which is an essential part of Optimization problems. It allows us to reach global minima faster without getting stuck at local minima. Additionally, EM-PSO having an additional tunable parameter i.e. exponentially averaged momentum adds flexibility to the task of exploration better than PSO or its vanilla momentum version. The computed weighted average in the Momentum particle swarm optimization (M-PSO) algorithm \cite{xiang2007pso} contributes to exploration and exploitation simultaneously. Locating the optima efficiently in the search space hinges on exploration for the better part, Therefore, it is reasonable to assign more weight to the exploration part of the PSO equation such that we benefit from greater weights. The momentum term in M-PSO also contributes to more iterations to reach the optima, as observed elsewhere \cite{saha2020adaswarn}. EM-PSO mitigates the issues faced by M-PSO and PSO by leveraging the exploration phase, determined by the \textbf{exponential weighted average of the historical velocities} only. The negligible weights in M-PSO do not aid the required acceleration. The momentum in EM-PSO is the exponential collection of velocities experienced by the particles over time. The velocities aggregate by an exponential multiplication factor $\beta$, as particles progress in time. Thus, $\beta$ factor is responsible for recent velocities having greater weights than their older counterparts.

\section{The Schrödinger Equation}
\label{sec:schrodinger}
\subsection{Introduction}

The time-dependent Schrödinger Equation is the central equation in non-relativistic quantum mechanics which defines the evolution of a particle $\psi(\va*{r}, t)$. It was postulated by Erwin Schrödinger in 1925 as a means to describe the behaviour of an electron \cite{griffiths} ---
\begin{gather}
    -\frac{\hbar^2}{2m}\laplacian\psi + V(\va*{r}, t)\psi = \iota\hbar\pdv{\psi}{t}
    \label{eq:time.dependent.schrodinger}
\end{gather}
Eq (\ref{eq:time.dependent.schrodinger}) is a partial differential equation involving spatial and temporal coordinates. Applying separation of variables on it \cite{griffiths}, one obtains the time-independent Schrödinger equation for $\psi(\va*{r})$ as follows ---
\begin{gather}
    -\frac{\hbar^2}{2m}\laplacian\psi + V(\va*{r})\psi = E\psi
    \label{eq:time.independent.schrodinger}
\end{gather}It is also an eigenvalue differential equation, and the potential function $V(\va*{r})$, along with appropriate boundary conditions, represents the physical situation in which the Schrödinger equation has to be solved. Taking eq (\ref{eq:time.independent.schrodinger}) in one dimension, eq (\ref{eq:uni.schrodinger}) is recovered. Putting $V(\va*{r})=0$, one can derive the interference phenomenon for the double-slit experiment with electrons, which is commonly cited as an example of wave-particle duality. The Schrödinger equation was conceived as the quantum equivalent of Newton's second law of motion. The mathematical prediction of the path of a given physical system over time as defined in classical mechanics is not adequate in the Quantum mechanics framework and requires the quantum-mechanical characterization of an isolated physical system, explained well by the Schrödinger equation. The evolution over time is represented by a wave function, under the assumptions of a unitary time-evolution operator generated by quantum Hamiltonian. The wave function, a complex-valued probability amplitude, is the quantum state of an isolated quantum system, described mathematically. 

\subsection{Applications in Quantum Mechanics}

All useful information about the system can be derived from the solution $\psi(\va*{r})$ such as its allowed energies (E), and angular momentum ($\va*{L}$), which is heavily applied in the theory of spectroscopy \cite{banwell}. However, there are only a few situations in which it could be solved exactly --- toy potentials such as particle in a box, delta function potential, and most notably the hydrogen atom --- otherwise it has to be approximated intelligently by physical symmetries/constraints \cite{bransden} of the problem, or solved numerically. For example, energy transitions in 1,3 butadiene \cite{butadiene} can be attributed to its alternating $C=C$ structure by treating the electrons in the chemical bonds as a particle-in-a-box problem. 

Solutions to the Schrödinger equation have immense applications in quantum chemistry, which seek to describe the behaviour of simple polyatomic molecules. Even in the elementary case of He$^{\text{4}}_2$ \cite{bransden}, the wavefunction is untenable to solve by Runge-Kutta methods, let alone analytically. Hence an alternative method to obtain approximate solutions to eq (\ref{eq:time.independent.schrodinger}) would benefit the present state of Quantum Chemistry and allow the study of tougher potentials $V(\va*{r})$.

\section{Using NN to solve Differential Equations}
\label{NN}

\subsection{Preliminaries}

A Neural Network (NN) is an artificially designed network that is inspired by the working of neurons in the human body and which is devised to find an association between data sets. Each neuron receives a column vector of features \(\boldsymbol{x}\) as input. The set of coefficients of \(\boldsymbol{x}\) are called as weights, denoted by \(\boldsymbol{w}\). The output \(\hat{y}\) is given by: \(\hat{y} = \sigma (z)\) where, \(\ z = \boldsymbol{w^T} \boldsymbol{x} + \epsilon\). Here, \(\sigma (z)\) is an activation function. In order to model complex figures to perform classification, many neurons are arranged in several layers. The Loss function represents the error of the current solution from the actual solution. In every iteration, the loss function is calculated to let the computer know how far it is from the ideal classification. After each iteration, the value of parameters \(\boldsymbol{w}\) need to be adjusted via a process of Back-propagation. The NN will try to reduce this loss in further iterations using a suitable back-propagation algorithm, which finds the minima of the loss function. This idea can be utilized to solve differential equations.
\subsection{Conversion to an Optimization problem}

A differential equation can be solved either analytically or by using numerical techniques. However, it is possible to use NNs to find a function that satisfies the differential equation. Since NN works on a back-propagation algorithm, it requires a cost function, which it will try to minimize. Thus, the differential equation can be converted to a minimization problem to resemble the cost function. Considering a \(2^{nd}\) order Boundary Value Problem (BVP) given by ---
\begin{gather}
    D[f] \equiv \dv[2]{f}{x} + a\dv{f}{x} + bf - c = 0 \label{eq:bvp} \\
    f(x_0) = f_0 \nonumber \\
    f(x_1) = f_1 \nonumber
\end{gather}Let the output of the NN be $u$. If we set ---
\begin{gather}
    \hat{u}=u_1\left(\frac{x-x_0}{x_1-x_0}\right) + u_0\left(\frac{x-x_1}{x_0-x_1}\right) + (x-x_0)(x-x_1)u \label{eq:bvp.nn}
\end{gather}We see that $\hat{u}$ automatically satisfies the boundary conditions $u(x_0)=u_0$ and $u(x_1)=u_1$ by construction. Hence if the NN could optimize the loss function $L = \left(D[\hat{u}]\right)^2$ identically to 0, the differential equation would be exactly solved and the boundary conditions obeyed. This construction (\emph{and similar others}) are described in the landmark paper by \emph{Lagaris et. al.} \cite{lagaris1998ann}, which also proposes methods to solve systems of ODEs/PDEs in higher dimensions.

\section{Our contribution}
\label{sec:contribution}
\subsection{Exponentially Averaged Momentum Particle Swarm Optimization (EM-PSO)}

\begin{algorithm}
\SetAlgoLined
\For{each particle p in Swarm S}{
initialize particle with feasible random position\;
evaluate the fitness \(F_i\) of the particle\;
}
\While{accuracy is within the desired limit}{
\For{each particle p in Swarm S}{
\(v_i =   M_i + c_1 r_1 (P_i - x_i) + c_2 r_2 (G - x_i) \)\;
\(x_i = x_i + v_i\)\;
\(M_i = \beta M_i + (1 - \beta) v_i\)\;
update \(P_i\) if fitness \(F_i\) has improved
}
update \(G\) if there is a new global best ;
}
\algorithmicreturn \ \(G\)
\caption{Exponentially Averaged Momentum Particle Swarm Optimization}
\end{algorithm}

The problem with the currently available Momentum Particle Swarm Optimization \cite{xiang2007pso} is that the computed weighted average takes care of both exploration and exploitation simultaneously. Since PSO tries to search the space by exploration, more weightage should be given to the exploration part. It also requires more iterations to reduce errors and reach the optimal value.
In this section, we explore a novel approach to Momentum Particle Swarm optimization, called Exponentially Averaged Momentum Particle Swarm Optimization \cite{mohapatra2021mpso}.  This model will try to address the above problems. It's characteristic iteration scheme is given by ---
\begin{align} 
M_i^{d+1} &= \beta M_i^d + (1 - \beta) v_i^d \label{eq:mom.recur}\\
v_i^{d+1} &=   M_i^{d+1} + c_1 r_1 (P_i^{d} - x_i^{d}) + c_2 r_2 (G^{d} - x_i^{d}) \\
x_i^{d+1} &= x_i^{d} + v_i^{d+1}
\label{empso.update.eqns}
\end{align}where \( x_i^d\) represents the position of particle \(i\) after \(d^{th}\) iteration, \(v_i^d\) is the velocity of particle \(i\) after \(d^{th}\) iteration, \(P_i\) represents the best position till now for the \(i^{th}\) particle, \(G\) is the best global position found till now, \(\beta\) is the momentum factor and \(M_i^{d+1}\) is the effect of momentum in \((d+1)^{th}\) iteration. Recursively expanding eq (\ref{eq:mom.recur}), we get ---
\begin{align}
    M_i^{d+1}  &= \sum_{k=0}^{d} \beta^k (1-\beta) v_i^{d-k} \nonumber & \label{solving_momentum}
\end{align}Since \(\beta < 1\), the Momentum is distributed in such a way that the focus is more on present velocities and lesser on the previous velocities (as the \(\beta^k\) factor piles up). As the iteration scheme progresses, the coefficient of the older velocities is piled up with the factor ($\beta$). This enhances the exploration power of PSO and also prevents particles from being stuck in local minima by weighted addition of its past velocities.

\noindent \textbf{Consequence of the Exponentially Averaged scheme:} The weight of the $(d-i)$th term is $\beta^i (1-\beta)$, with $\beta$ restricted to less than 1. Such terms grow smaller when $\beta$ is exponentiated with a positive number. This results in older
velocities being assigned weights and, therefore, they contribute less to the overall value of the Momentum. Since the velocities are a cumulative sum, no additional memory is required to keep historical velocities.

\subsection{Complexity Analysis of EM-PSO}
\label{order}

For determining the complexity of an algorithm, we need to count the number of primitive operations. 
\begin{itemize}
    \item Steps 1-3 consume a fixed number of operations for the evaluation of a swarm on any objective function and hence, can be excluded. Generally, the swarm size is fixed to 25/50/100, and hence the time cost incurred in initialization can be ignored when compared to that incurred in the evaluation of the swarm until covergence.
    \item With respect to the guard of the while loop at line 5, assume the swarm does not terminate upto $t$ iterations.
    \begin{itemize}
        \item It takes $\bigoh{1}$ time to update the position, velocity, momentum, fitness (required for personal best update) of a particle.
    \end{itemize}
    \item There are $n$ particles and it takes $\bigoh{n}$ time to update the global best per iteration, if necessary.
\end{itemize}
This gives an overall time complexity of $\bigoh{nt}$ for EMPSO.

\subsection{Stability Analysis of EM-PSO}
\label{stability}

A deterministic version of EM-PSO (\emph{fixed pbest and gbest}) could be looked at as a finite difference scheme in $x$ and $v$. In this light, we prove its Von-Neumann stability \cite{some.math.textbook} which accounts for the stability of EM-PSO.

\noindent \textbf{Theorem:} Exponentially Averaged Momentum Particle Swarm Optimization with the momentum factor $\beta$ is said to be stable iff the acceleration coefficients \(c_1, c_2\) and $\beta$ satisfy the conditions:
\begin{itemize}
    \item $0 < \beta <1$
    \item $0 \leq (c_1+c_2) \leq 2$
\end{itemize}

\noindent \textbf{Proof:} In Section IV, we defined the velocity update rule as follows, 
\begin{equation}
 v_i^{d+1} =   M_i^{d+1} + c_1 r_1 (P_i^{d} - x_i^{d}) + c_2 r_2 (G^{d} - x_i^{d})
\end{equation}where,
$c_1 = $ weight of the local information $\times r_1$, $c_2 = $ weight of global information $\times r_2$. Assuming iteration process is near global minima, we consider \(P_i^{d} = p_1\) and \(G^{d} = p_2\).
Combining with the position update rule, we have,\begin{equation}
   x_i^{d+1} = x_i^{d} + M_i^{d+1} + c_1(p_1 - x_i^{d}) + c_2(p_2 - x_i^{d}) 
\end{equation}

Expansion of $ M_i^{d+1}$, followed by repeated substitution of $v^{d+1}=x^{d+1} - x^d$, we get ---
\begin{equation}
\label{eqn:finitediff}
\begin{split}
    x_i^{d+1} = (2- \beta - c_1 - c_2) x_i^{d} - (\beta^2 -2\beta+1) x_i^{d-1}\\
    - \beta(1-\beta) x_i^{d-2} + c_1 p_1 + c_2 p_2
\end{split}
\end{equation}

Eq (\ref{eqn:finitediff}) is our desired difference scheme. We apply the transformation $d \rightarrow d+2$, along with setting $c_1p_1 + c_2p_2=0$ to obtain the corresponding homogeneous scheme ---
\begin{equation}
\label{eqn:finitenewscheme}
        x_i^{d+3} - \lambda_1  x_i^{d+2} + \lambda_2  x_i^{d+1}
    - \lambda_3  x_i^{d} = 0
\end{equation}where,
\begin{itemize}
    \item $\lambda_1 = 2- \beta - c_1 - c_2$
    \item $\lambda_2 = \beta^2 -2\beta+1$
    \item $\lambda_3 = \beta(1-\beta)$
\end{itemize}

Its characteristic equation is 
\begin{equation}
    A^3 - \lambda_1 A^2 + \lambda_2 A - \lambda_3 = 0
\end{equation}where,  $A = $ Amplification Factor. Now, according to von Neumann's stability criterion, the finite differences scheme \eqref{eqn:finitenewscheme} is stable iff for the amplification factor (A), $|A| \leq 1$. Therefore, the finite differences scheme given by equation \eqref{eqn:finitenewscheme} and by equation \eqref{eqn:finitediff} ensure that the EM-PSO algorithm is stable iff $|A| \leq 1$. On finding the root of the cubic equation, we conclusively say that EM-PSO is stable iff ---
\begin{itemize}
    \item $0 < \beta <1$
    \item $0 \leq (c_1+c_2) \leq 2$
\end{itemize}
\noindent \textbf{Consequence of the Stability Analysis:} The adoption of hyper-parameters $\beta, c_1, c_2$ are facilitated by the analysis as it helps constrict the search space for optimization. We choose $\beta=0.9, c_1=0.8, c_2=0.9$ guided by EM-PSO fundamentals and stability analysis results.

\subsection{Working of EM-PSO to replace back-propagation}

Suppose, initially there is a NN given by, \(y = N_a(\boldsymbol{x})\), which has a total of \(n\) neurons arranged in the form of layers, with \(k\) weights in the entirety of the network. Hence, \(y = f(x,w_1, w_2, w_3, ... , w_k)\). Clearly, in order to find \(y\) based on the data, the optimal value of the weights \(\boldsymbol{w}\) must be found. Hence, it is possible to establish an equivalence between the weights \(\boldsymbol{w}\) and the global best position \(G\) i.e. the dimension over which EM-PSO is run is equal to the number of weights (here, EM-PSO would be run in space \(R^k\) ) and each component \(G_i\) of \(G = (G_1, G_2, G_3, ...G_i,.. G_k)\) represents a weight \(w_i\). We can include hidden layers as well, which would add more weights to our NN model and consequently, add more dimensions to the space of the particle swarm framework.

The forward propagation in \(N_a\) is when the value \(y\) is computed as \(y = N_a(\boldsymbol{w}, x)\), based on previous iteration weights. Meanwhile, one iteration of EM-PSO will mean that a better approximation for the global solution is found \(\Rightarrow\) a better approximation is found for \(\boldsymbol{w}\). Hence, EM-PSO acts like a back-propagation algorithm in finding the right weights. 

While considering a working model, since the domain \(X\) will be discretized into \(m\) points i.e. \(\bar{y} = N_a(\boldsymbol{w}, \bar{x})\), the differentials can be expressed as:
\begin{align*}
\bar{dy} &= \delta N_a (\boldsymbol{w},\bar{x}) \\ 
\bar{d^2y} &= \delta \bar{dy} = \delta^2 N_a (\boldsymbol{w},\bar{x})
\end{align*}The cost function, being a function of differentials, can now be computed.

\subsection{Using EM-PSO to Solve Particle-in-a-Box}
\label{5.5}
\begin{figure}[h]
    \centering
    \includegraphics[width=75mm]{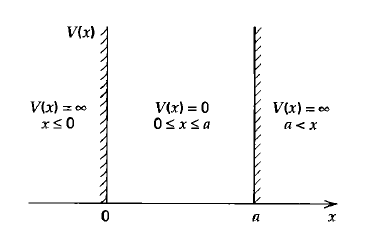}
    \caption{Box Potential}
    \label{fig:box.potential}
\end{figure}

In this paper, we restrict ourselves to the \emph{Particle in a Box} potential for the time-independent Schrödinger equation (\textit{L is the length of the box}) -
\begin{gather}
V(x)=
\left\{
    \begin{array}{ll}
        0  & ; \ 0 \leq x \leq a \\
        \infty & ; \ \text{otherwise}
    \end{array}
\right. \label{eq:piab.potential}
\end{gather}

The infinite potential kills the wavefunction outside the box to $\psi(x)=0$. The analytical solution inside $0 \leq x \leq a$ is ---
\begin{gather}
\psi(x) = \sqrt{\frac{2}{a}} sin\left(\frac{n\pi x}{a}\right) \label{eq:piab.sol}
\end{gather}

with the eigenenergy ---
\begin{gather}
E_n = \frac{n^2 \hbar^2 \pi^2}{2 m a^2} \label{eq:piab.eigen}
\end{gather}

and the boundary conditions ---
\begin{gather}
    \psi(0) = \psi(a) = 0 \label{eq:piab.boundary}
\end{gather}

For the sake of convenience, we set $\hbar=m=a=1$ for our neural network training. The eigenenergy $E$ also needs to be added as a learnable parameter to the neural network, which EM-PSO will optimize. It was observed in preliminary experiments that the network had a tendency to learn the trivial function $\psi(x)=0$. This physically violates the wavefunction normalization rule of QM \cite{griffiths} ---
\begin{gather}
    p \equiv \int_{-\infty}^{\infty} |\psi(x)|^2 \dd x = 1 \label{eq:wvfn.norm}
\end{gather}

Its physical interpretation is that the particle has a full probability of being found in all of space. To enforce the neural network to learn a non-trivial $\psi(x)$, we add \emph{probability regularization terms} ($p=\int_0^1 |\psi(x)|^2 \dd x$) to the loss function $L$. This is similar to the inverse regularization terms used in \cite{quantum.unsupervised}, which drives the network to learn a physically useful solution.
\begin{equation} \label{eq:prob.reg}
\begin{split}
R(p) = (1-p)^2 + \frac{20}{p} + \frac{5}{p^2} + \frac{5}{6p^3} + \frac{5}{48p^3}\\
+ 20p + 5p^2 + \frac{5p^3}{6} + \frac{5p^4}{48} 
\end{split}
\end{equation}

\begin{figure}[h]
    \centering
    \includegraphics[width=90mm]{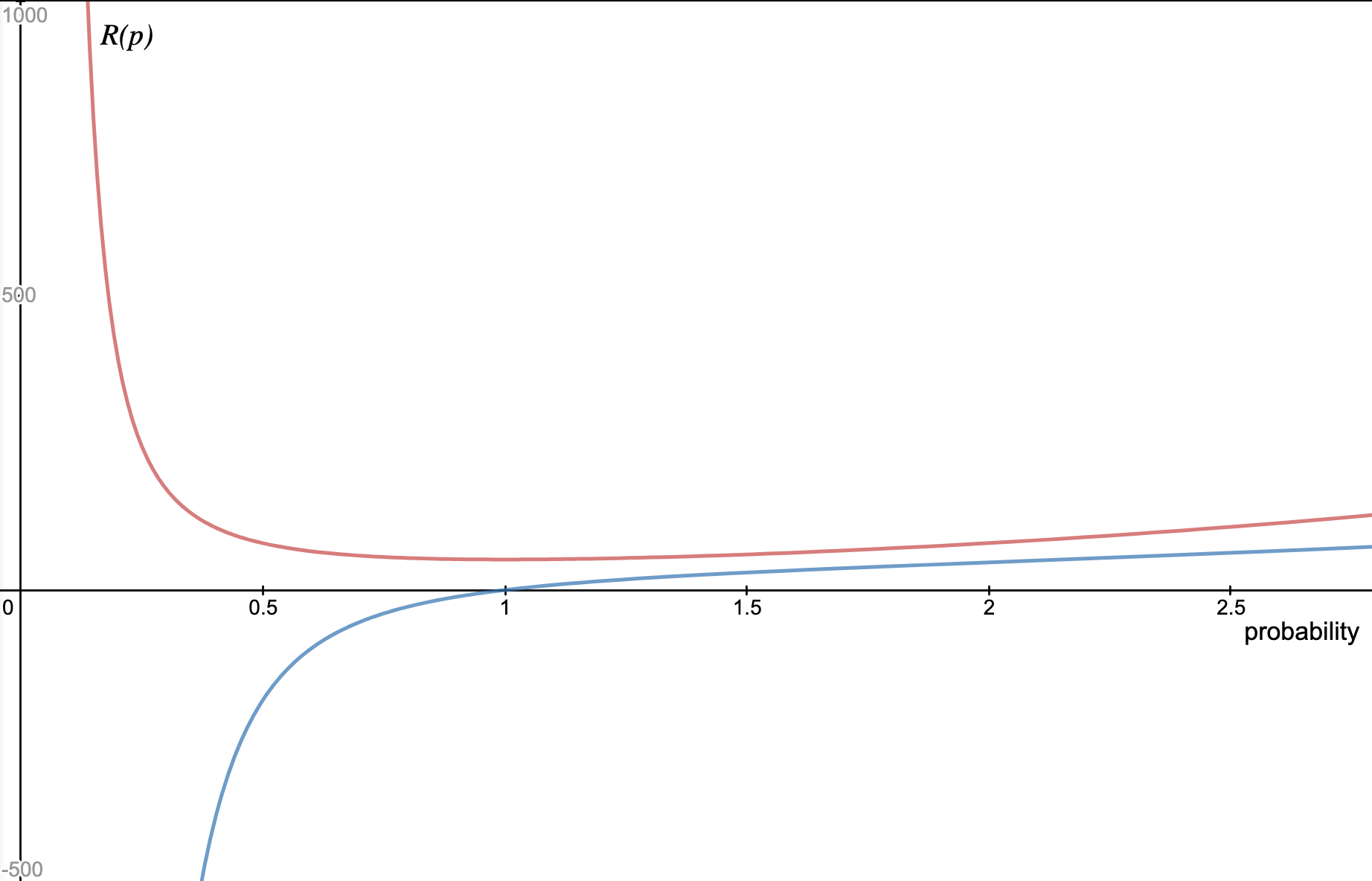}
    \caption{Regularization Function (Red $\rightarrow R(p)$, Blue $\rightarrow \dv{R}{p}$)}
    \label{fig:probreg}
\end{figure}

The $(1-p)^2$ term drives the probability to unity as $p=1$ is a global minimum for $(1-p)^2$. The inverse terms $\frac{1}{p^\alpha}$ terms prevent the probability from vanishing, and the $p^\alpha$ terms prevent it from exploding. The choice of the coefficients is such that the following relation is satisfied ---
\begin{gather}
    \delta_R(p^\alpha) = 2\delta_R(p^{\alpha+1}) \label{eq:prob.rel1} \\
    \delta_R \left(\frac{1}{p^\alpha}\right) = 2\delta_R \left(\frac{1}{p^{\alpha+1}}\right) \label{eq:prob.rel2}
\end{gather}

where $\delta_R$ is the contribution of each regularization term to the loss function via $R(p)$. We justify our choice of constraints eq (\ref{eq:prob.rel1}, \ref{eq:prob.rel2}) in the sense that regularization term with exponent $\alpha + 1$ (\emph{inverse or polynomial}) contributes half of that of $\alpha$, to the gradient. Our goal is to ultimately optimize the loss $\left(D[\hat{u}]\right)^2$ (eq \ref{eq:bvp}) and it would be undesirable to have the regularization terms dominate the gradient of the neural network. Setting the coefficients of $p$ and $\frac{1}{p}$ as $20$, the coefficients for the higher exponents naturally follow and can be derived by considering the expression for $\dv{R}{p}$. The plot of the regularization function $R(p)$ is informative (figure \ref{fig:probreg})

\subsubsection{Challenges in Computation}

The primary bottleneck in the optimization process is the evaluation of the integrated probability density over the entire spatial domain. Moreover, the computation of the loss function at each iteration of EM-PSO involves $\dv[2]{\psi}{x}$ (\emph{and } $\dv{\psi}{x}$ \emph{as an intermediate step}),  which is an expensive operation for automatic differentiation.

\section{Results}
\label{results}

\begin{figure}[h]
     \centering
     \begin{subfigure}[b]{0.45\textwidth}
         \centering
         \includegraphics[width=\textwidth]{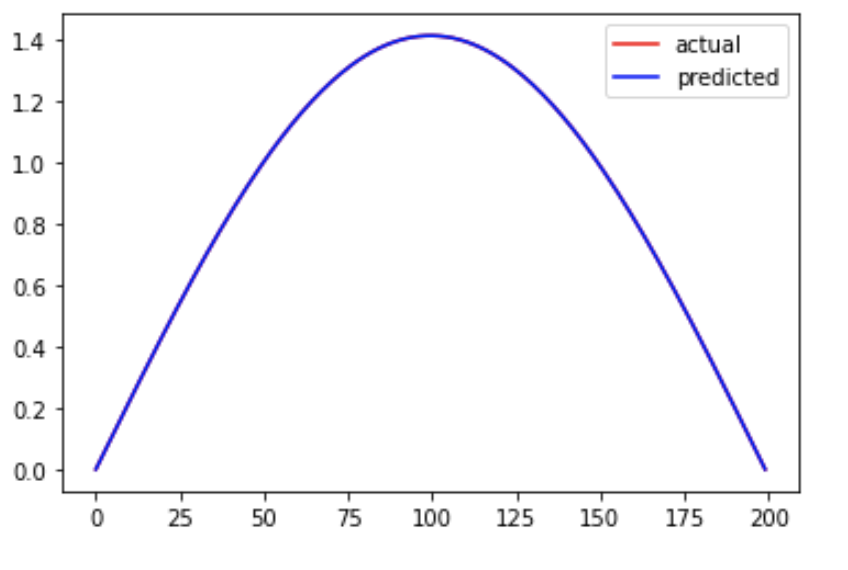}
         \caption{Wavefunction}
         \label{fig:results.wavefunction(n=1,l=1)}
     \end{subfigure}
     \begin{subfigure}[b]{0.45\textwidth}
         \centering
         \includegraphics[width=\textwidth]{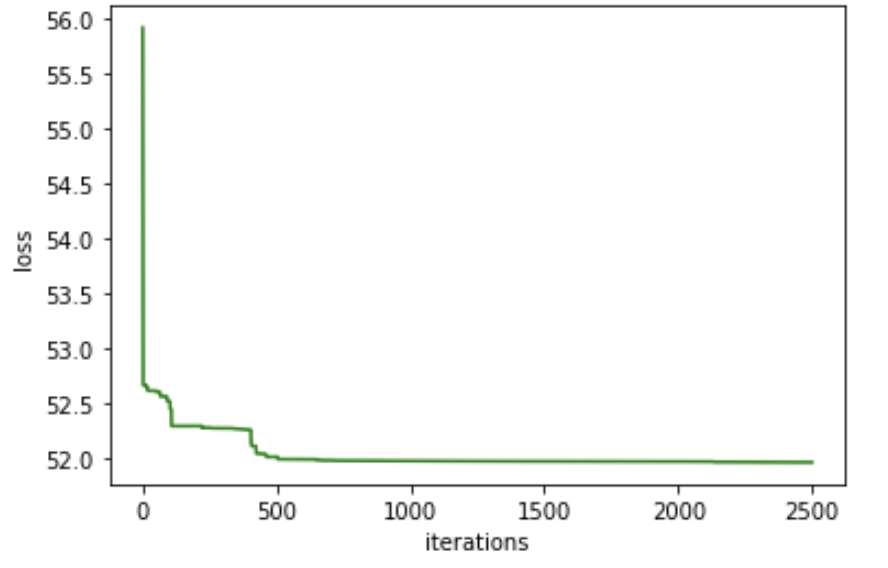}
         \caption{Total Loss vs. Iterations, total loss includes MSE of $D[f]$ and regularisation function value, minimum total loss is 51.875 in our case due to the regularisation function.(Here we are showing only first 2500 steps)}
         \label{fig:results.losses(n=1,l=1)}
     \end{subfigure}
     \caption{Plots for Neural network trained for n=1, l=1}
\end{figure}

We have obtained results for the Particle-in-a-Box potential with $n = 1 , l = 1$. The eigenvalue for this problem is 4.9348022(i.e.$\frac{\pi^2}{2}$) and our model learned eigenvalue as 4.9346618 with random initialization of E in the range of [4,6] in 5000 epochs. The learnt wave function and loss curve are in figures (\ref{fig:results.wavefunction(n=1,l=1)}, \ref{fig:results.losses(n=1,l=1)}) respectively.

\begin{figure}[h]
     \centering
     \begin{subfigure}[b]{0.45\textwidth}
         \centering
         \includegraphics[width=\textwidth]{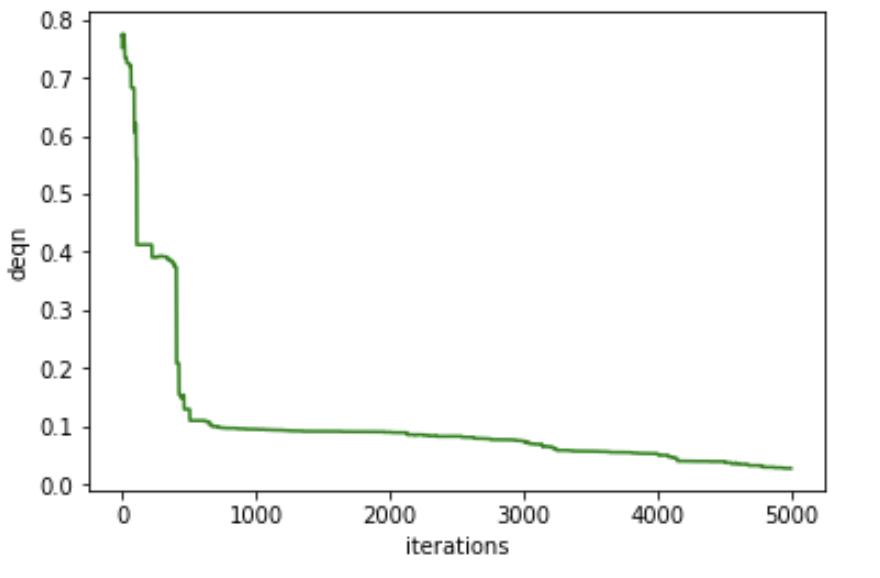}
        \caption{MSE of $D[f]$ vs. Iterations, $D[f]=0$ is the equation. For neural network to imitate solution of differential equation, $D[f]$ should be zero at all the points in domain. So we use it in the loss function.}
        \label{fig:results.deqn(n=1,l=1)}
     \end{subfigure}
     \begin{subfigure}[b]{0.45\textwidth}
         \centering
        \includegraphics[width=\textwidth]{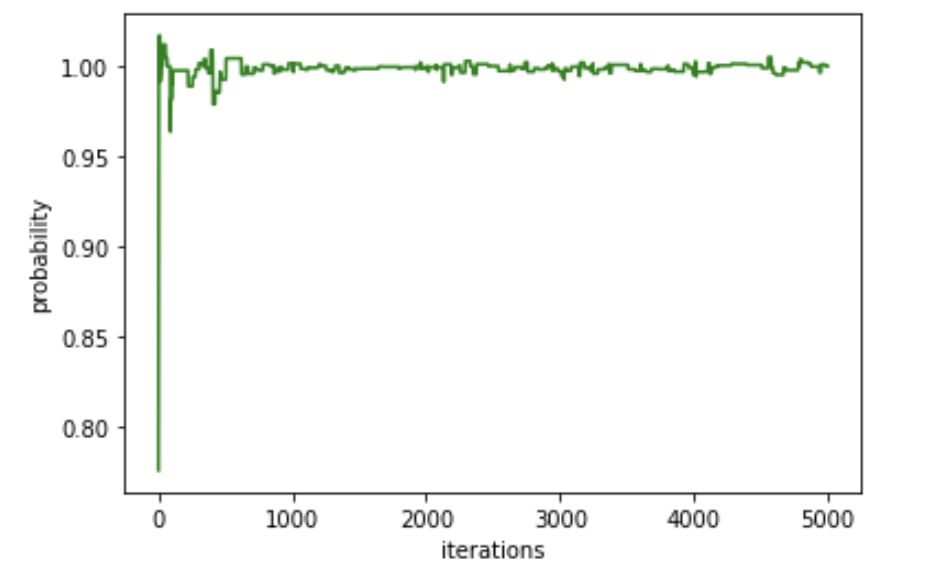}
        \caption{Integrated probability vs. Iterations, regularisation with respect to probability is important otherwise it learns trivial solution $f=0$, so it is important to monitor probability}
        \label{fig:results.probs(n=1,l=1)}
     \end{subfigure}
     \caption{Plots for Neural network trained for n=1, l=1}
\end{figure}

Some of the important values learnt by the model at the end of 5000 epochs are \emph{loss = 51.8908180} (includes differential equation error and probability regularization function) The integrated loss of $D[f]$ over $[0,1] = 0.0158156$ and integrated probability density = $1.0000519$. Graphs/plots for this experiment are shown in figures (\ref{fig:results.wavefunction(n=1,l=1)}, \ref{fig:results.losses(n=1,l=1)}, \ref{fig:results.deqn(n=1,l=1)}, \ref{fig:results.probs(n=1,l=1)}). It is also insightful to note that we had also obtained \emph{out-of-phase} solutions to the particle-in-a-box in our experiments (figure \ref{fig:results.wavefunction(outOfPhase)(n=1,l=1)}). This is because eq (\ref{eq:time.independent.schrodinger}) is invariant to a change of sign $\psi(x) \rightarrow -\psi(x)$. Physically, this means that the particle has no preferential direction of motion in the x-coordinate \cite{griffiths}. 
 
\begin{figure}[H]
    \centering
    \includegraphics[width=75mm]{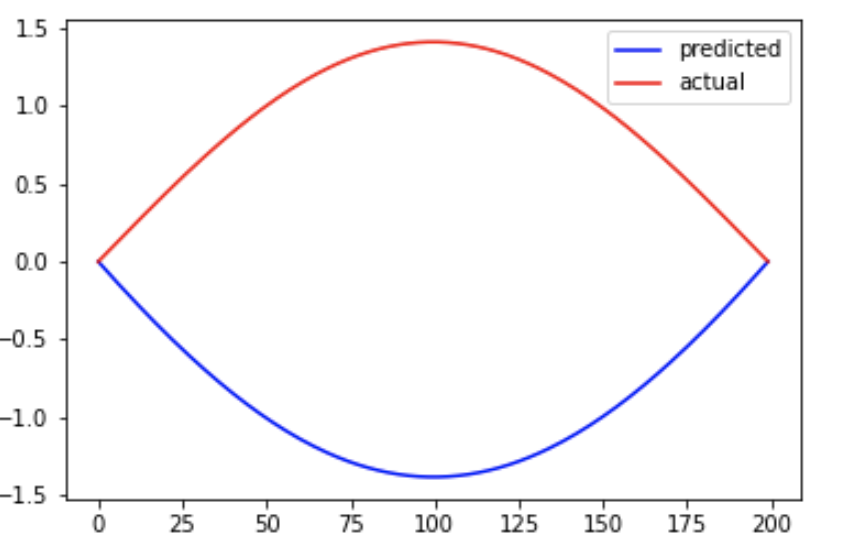}
    \caption{Wavefunction(out phase) obtained in one of our experiments(n=1, l=1)}
    \label{fig:results.wavefunction(outOfPhase)(n=1,l=1)}
\end{figure}

Similarly, we also obtained results for potential with $n = 2 , l = 1$. Eigenvalue for this problem is 19.7392088(i.e.${2}{\pi^2}$) and our model learned eigenvalue as 19.7383102 with random initialisation of E in the range of [19,21] in 10000 epochs. The learnt wavefunction and loss curve are in figures (\ref{fig:results.wavefunction(n=2,l=1)}, \ref{fig:results.losses(n=2,l=1)}) respectively. Some of the important values learnt by the model at the end of 10000 epochs are \emph{loss = 51.9017940} (includes differential equation error and probability regularization function) The integrated loss of $D[f]$ over $[0,1] = 0.0160273$ and integrated probability density = $1.0001679$. Graphs/plots for this experiment are shown in figures (\ref{fig:results.wavefunction(n=2,l=1)}, \ref{fig:results.losses(n=2,l=1)}, \ref{fig:results.deqn(n=2,l=1)}, \ref{fig:results.probs(n=2,l=1)}). 

\begin{figure}[h]
     \centering
     \begin{subfigure}[b]{0.45\textwidth}
         \centering
         \includegraphics[width=\textwidth]{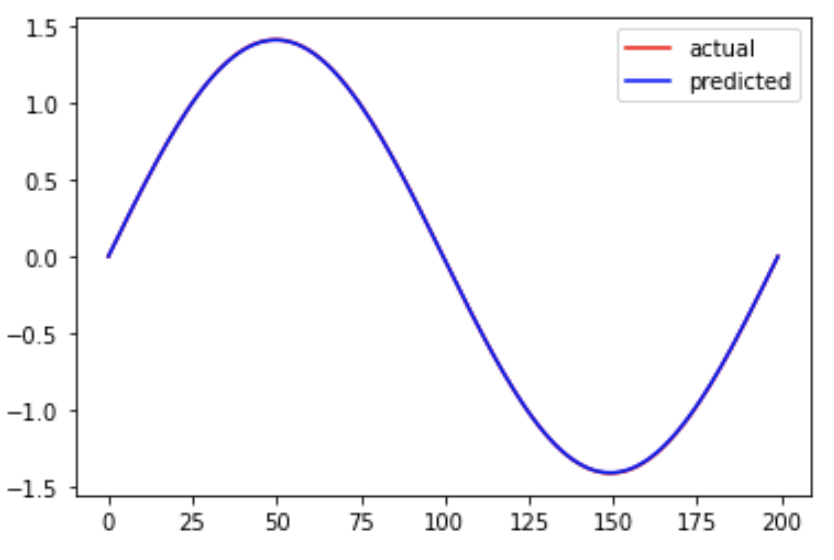}
         \caption{Wavefunction}
         \label{fig:results.wavefunction(n=2,l=1)}
     \end{subfigure}
     \begin{subfigure}[b]{0.45\textwidth}
         \centering
         \includegraphics[width=\textwidth]{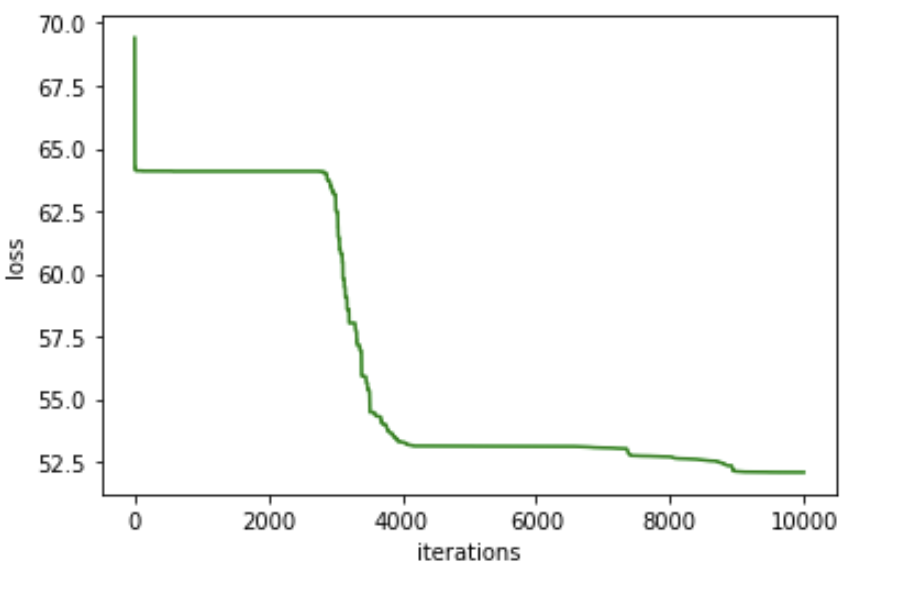}
         \caption{Total Loss vs. Iterations, total loss includes MSE of $D[f]$ and regularisation function value, minimum total loss is 51.875 in our case due to the regularisation function.}
         \label{fig:results.losses(n=2,l=1)}
     \end{subfigure}
     \caption{Plots for Neural network trained for n=2, l=1}
\end{figure}

\begin{figure}[h]
     \centering
     \begin{subfigure}[b]{0.45\textwidth}
         \centering
         \includegraphics[width=\textwidth]{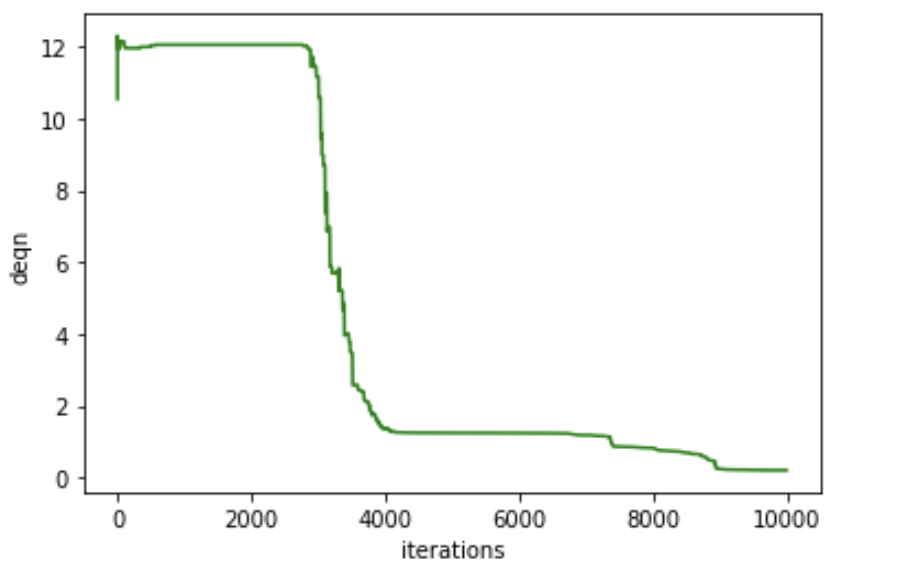}
        \caption{MSE of $D[f]$ vs. Iterations, $D[f]=0$ is the equation. For neural network to imitate solution of differential equation, $D[f]$ should be zero at all the points in domain. So we use it in the loss function.}
        \label{fig:results.deqn(n=2,l=1)}
     \end{subfigure}
     \begin{subfigure}[b]{0.45\textwidth}
         \centering
        \includegraphics[width=\textwidth]{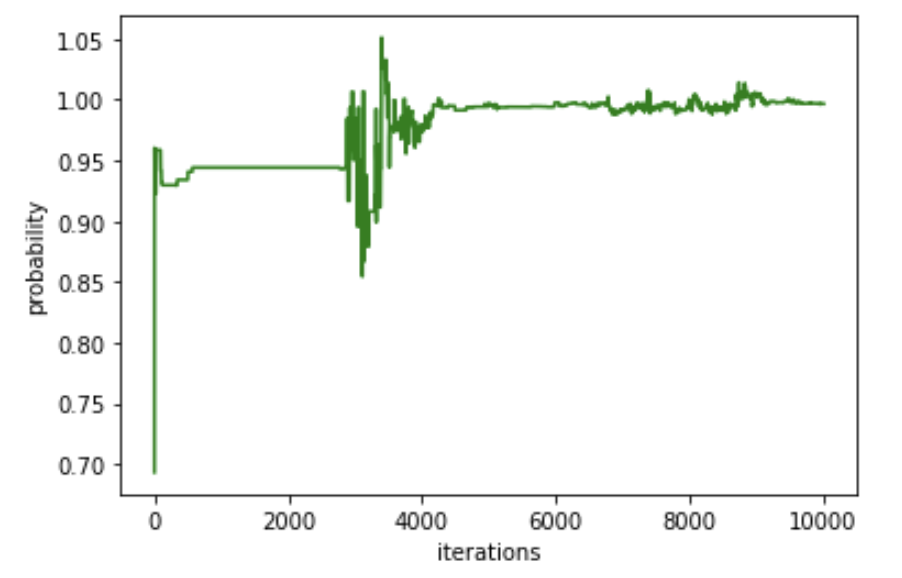}
        \caption{Integrated probability vs. Iterations, regularisation with respect to probability is important otherwise it learns trivial solution $f=0$, so it is important to monitor probability}
        \label{fig:results.probs(n=2,l=1)}
     \end{subfigure}
     \caption{Plots for Neural network trained for n=2, l=1}
\end{figure}

The functions $\psi_n(x)=\sqrt{\frac{2}{a}}\sin \left(\frac{n\pi x}{a}\right)$ represent a set of basis eigenfunctions to the particle-in-a-box potential with eigenenergy $E_n$. It is precisely in the form of an \emph{eigenvalue differential equation} that we have obtained solutions $\langle\psi_n(x), E_n\rangle$ for various $n$. The complete solution to the \emph{time-dependent} equation for the particle-in-a-box would be ---
\begin{gather}
\Psi_n(x,t) = \psi_n(x) e^{-\frac{\iota E_n t}{\hbar}}
\end{gather}
The $e^{-\frac{\iota E_n t}{\hbar}}$ had been extracted out by separation of variables (section \ref{sec:schrodinger}) and details can be found in \cite{griffiths}. Any linear combination  $\Psi(x,t) = \sum_n c_n \psi_n(x)e^{-\frac{\iota E_n t}{\hbar}}$ would also be a solution to the \emph{time-dependent} Schrödinger Equation owing to the fact it is a linear partial differential equation. What we have \emph{not} done in our work is to solve the equation in the general sense where solution may not be a stationary state (\emph{eigenfunction}). It is precisely for stationary states (\emph{stationary in time}) that the \emph{time-independent} Schrödinger equation can be formulated and solved, which we have demonstrated for the particle-in-a-box. For the general case of the time-dependent problem, the techniques of \cite{lagaris1998ann} would need to be employed to solve the problem in the NN framework.

\section{Summary and Conclusion}
\label{summary}

We have introduced an application of the meta-heuristic algorithm EM-PSO, which is a variant of Vanilla PSO. It is inspired by existing works that solve differential equations by NN. However, no approach to solve Differential equations using metaheuristics driven PSO has ever been attempted. Subsequently, we developed EM-PSO in such a way that it enhances the exploration power of PSO and prevents the particles from being stuck in the local minima by using an exponentially weighted addition of its past velocities, and discussed the importance of Schrödinger Equation and it's applications in Quantum Mechanics. Moreover, we have provided the methodology to convert an ordinary differential equation to an Optimization problem and presented the framework of EM-PSO to replace backpropagation in a neural network. Finally, we analysed important aspects of this algorithm such as the Stability and Complexity, supported by concrete mathematical proofs, and applied this method to the Particle-in-a-Box problem with satisfactory results.

\par Due to the lack of control in the eigenvalue derivatives, it could be possible that it crosses any of the singularities of the regularization terms, or lands exactly on it. There is no recovery from this situation in the present framework. Particle in a Box is further simplified by the fact that the potential is infinity outside the box. We know \textit{a priori} that the wave function vanishes in such regions. However, for the generic potential, the square integrability of the wave function has to be checked for $x \xrightarrow{} \pm \infty$. Using any numerical integration technique for the entire position space could slow down the training immensely, and alternate faster integration techniques need to be sought.
\par Our future plan is to conceive of architecture of two simultaneous NN - \textit{one for the eigenfunction, and the other for the eigenvalue} - that is trained over a common loss function. It has been noticed that the eigenvalue changes in similar magnitudes irrespective of the scale of the system (L), which is clearly a problem. This could be overcome by having a separate neural network output the eigenvalue. In some sense, the nodes backing the eigenvalue output offer more control over its change, and it could dynamically take care of the scale of the problem. In the case of eigenvalue regularization, \cite{adam.kingma} could be modified to externally increase eigenvalue derivative dynamically if it is noticed that it has spent many too many epochs near either of the singularity points.
\par We also plan to introduce a higher-order version of EM-PSO, which will use Hessian matrices for second-order approximations of the gradients. This will ensure more stability, faster convergence to the global minima, and reliable results in highly non-convex loss functions. 

\section*{Acknowledgement}
The authors would like to thank the Science and Engineering research Board (SERB), Department of Science and Technology, Government of India, for supporting our research by providing us with resources to conduct our experiments. The project reference number is: EMR/2016/005687.

\bibliographystyle{IEEEtran}
\bibliography{references}

\end{document}